\documentclass{article}

\usepackage{arxiv}
\usepackage{lineno}
\usepackage{amsthm}
\usepackage{graphicx}
\usepackage{amsmath}
\usepackage{amssymb}
\usepackage{indentfirst}
\usepackage{setspace}
\usepackage{boondox-cal}
\usepackage{enumerate}
\usepackage{microtype}
\usepackage{xcolor}
\usepackage{hyperref}
\definecolor{deepgreen}{rgb}{0,0.5,0}
\hypersetup{
    colorlinks=true,
    citecolor={deepgreen}
}
\usepackage{color}
\usepackage{soul}
\usepackage{subcaption}
\usepackage{array}
\usepackage{multirow}

\newcolumntype{C}[1]{>{\centering\let\newline\\\arraybackslash\hspace{0pt}}m{#1}}

\newtheorem{theorem}{Theorem}
\newtheorem*{problem}{Problem}

\title{Deep Learning-Based Quasi-Conformal Surface Registration for Partial 3D Faces Applied to Facial Recognition}
\date{}

\author{ Yuchen Guo \\
	Department of Mathematics\\
	The Chinese University of Hong Kong\\
	\And
        Hanqun Cao\\
	Department of Mathematics\\
	The Chinese University of Hong Kong\\
	\And
        Lok Ming Lui \\
	Department of Mathematics\\
	The Chinese University of Hong Kong\\
}

\renewcommand{\shorttitle}{\textit{arXiv} Preprint}

\begin{document}

\maketitle
\begin{abstract}
3D face registration is an important process in which a 3D face model is aligned and mapped to a template face. However, the task of 3D face registration becomes particularly challenging when dealing with partial face data, where only limited facial information is available. To address this challenge, this paper presents a novel deep learning-based approach that combines quasi-conformal geometry with deep neural networks for partial face registration. The proposed framework begins with a Landmark Detection Network that utilizes curvature information to detect the presence of facial features and estimate their corresponding coordinates. These facial landmark features serve as essential guidance for the registration process. To establish a dense correspondence between the partial face and the template surface, a registration network based on quasiconformal theories is employed. The registration network establishes a bijective quasiconformal surface mapping aligning corresponding partial faces based on detected landmarks and curvature values. It consists of the Coefficients Prediction Network, which outputs the optimal Beltrami coefficient representing the surface mapping. The Beltrami coefficient quantifies the local geometric distortion of the mapping. By controlling the magnitude of the Beltrami coefficient through a suitable activation function, the bijectivity and geometric distortion of the mapping can be controlled. The Beltrami coefficient is then fed into the Beltrami solver network to reconstruct the corresponding mapping. The surface registration enables the acquisition of corresponding regions and the establishment of point-wise correspondence between different partial faces, facilitating precise shape comparison through the evaluation of point-wise geometric differences at these corresponding regions. Experimental results demonstrate the effectiveness of the proposed method.
\end{abstract}


\section{Introduction}

% Facial landmark detection and registration are two key tools for conducting investigations on human faces like recognition or classification. In computer vision, the problems are widely investigated both in 2D and 3D, and researchers have proposed various methods to tackle them while getting the entire face of a person. However, in the actual scenario, it is difficult to acquire a full human face all the time since there could usually be things hiding the mouth or eyes. Thus, it is crucial to investigate the facial landmark detection on partial faces and classify the faces by a few overlapping.

%Facial landmark detection and registration play a crucial role in various applications involving human faces, such as face recognition, facial expression analysis, and face attribute classification~\cite{khabarlak2022fast,wu2019facial,sariyanidi2014automatic}. 
The shape analysis of the human face has garnered significant interest among researchers, emerging as a captivating and widely explored field of study\cite{marin2021spectral, brunton2014review, de2011facial}. It holds significant importance across diverse domains such as computer vision, computer graphics, and medical imaging, where it finds extensive applications. In computer vision, facial shape analysis plays a pivotal role in tasks such as facial recognition and facial expression analysis\cite{taskiran2020face, meng2021magface, khabarlak2022fast,wu2019facial,sariyanidi2014automatic}. Similarly, in the realm of medical imaging, the analysis of facial shape proves invaluable in medical diagnosis and treatment planning\cite{chan2020quasi}.

2D facial analysis has been extensively studied and is the subject of a surplus of literature. Nevertheless, recent advancements in 3D acquisition devices have revolutionized the field by facilitating the availability of rich 3D data of human faces. This surpasses the limitations of 2D images by providing a wealth of geometric information. Notably, numerous mathematical models for 3D facial analysis have been developed, yielding promising results\cite{hu2020face, lattas2020avatarme, moreano2020global}. A critical step in conducting shape analysis of 3D faces involves establishing accurate and dense pointwise correspondence among different human faces. This dense correspondence permits systematic and quantitative comparison, enabling comprehensive shape analysis\cite{qiu2020inconsistent, zeng2014surface}. While 3D surface registration of complete faces has been extensively studied, real-world scenarios often present challenges such as occlusions, partial view angles, or self-occlusions caused by hands or accessories. When faced with limited data availability, achieving correspondence between 3D partial faces becomes a more intricate task, commonly referred to as the {\it partial face registration} problem. The challenge lies in accurately determining the corresponding regions between two partially observed faces and establishing point-wise correspondence between these regions\cite{zhang2015learning, chen2021deep,guo2022automatic}. This particular class of surface registration problem is comparatively less studied. Without precise registration between partial faces, shape analysis and comparison among them cannot be performed accurately.

In this paper, we introduce a novel deep learning-based method that addresses the challenge of partial face registration. Our approach combines quasiconformal theories and deep neural networks to achieve accurate landmark detection and surface registration, even when dealing with partial facial data. This method can be further applied to facial recognition tasks. Our framework begins with a Landmark Detection Network (LD-Net), which precisely detects landmarks on the partial face surfaces using the curvature values derived from each vertex. This innovative approach allows us to accurately locate facial landmarks, even in scenarios with limited facial information or occlusions. These detected facial landmarks play a crucial role in guiding the subsequent registration process.
To establish a dense correspondence between the partial face and the template surface, we employ a registration network based on quasiconformal theories. This network consists of the Coefficients Prediction Network (CP-Net), which outputs the optimal Beltrami coefficient representing the surface mapping. The Beltrami coefficient quantifies the local geometric distortion of the mapping. By controlling the magnitude of the Beltrami coefficient through a suitable activation function, we can effectively control the bijectivity and geometric distortion of the mapping. The optimal Beltrami coefficient is then fed into the Beltrami solver network to reconstruct the corresponding mapping. The surface registration achieved by our method enables the acquisition of corresponding regions and the establishment of point-wise correspondence between different partial faces. This facilitates precise shape comparison by evaluating point-wise geometric differences at these corresponding regions. To thoroughly evaluate the effectiveness of our proposed method, we conducted extensive experiments on partial facial surfaces. The results obtained from these experiments  provide strong evidence for the robustness and reliability of our approach in achieving accurate landmark detection and surface registration for partial faces. The results of facial classification using the partial face registration further demonstrate the effectiveness of our proposed framework for analyzing partial faces.

In summary, the contribution of our work is as follows:

\begin{enumerate}[(I)]
    
    \item Automatic extraction of feature landmarks on partial faces: We propose a deep neural network that automatically extracts feature landmarks on partial faces by leveraging surface curvature information. The network accurately identifies their presences and precise locations, thereby facilitating subsequent registration.
    
    \item Quasi-conformal deep neural network for 3D partial face registration: We propose a novel approach that combines quasi-conformal (QC) geometry with convolutional neural networks (CNNs) to address the challenge of registering partial face surfaces. By integrating these techniques, we can improve both the accuracy and efficiency of the registration process, even in the presence of occlusions or limited facial information.

    \item Facial recognition of partial faces: By achieving precise surface registration of partial faces, we can identify the overlapping regions on each partial face and establish point-wise correspondences between them. This facilitates shape comparison and enables the recognition of partial faces through the evaluation of geometric differences at key feature landmarks.

   % \item Real-time algorithm for Quasi-conformal mapping generation:  To achieve precise surface registration with minimal distortion, we develop a real-time algorithm that utilizes the extracted landmarks and curvature values to generate quasi-conformal mappings. By leveraging the benefits of quasi-conformal geometry, our approach enables accurate registration even in the presence of occlusions or limited facial information.

    % \item We apply quasi-conformal geometry and CNNs for partial face surface registration.

    % \item Feature landmarks can be automatically extracted from the given partial faces based on the surface curvature.

    % \item We develop a real-time algorithm that generates quasi-conformal mappings using the extracted landmarks and curvature values, enabling precise surface registration with minimal distortion.

    % \item Our approach achieves satisfactory results in face recognition tasks, even when dealing with occlusions or limited facial information, enhancing the robustness of face classification systems.
    
    %\item Quasi-conformal mappings can be automatically generated based on the landmark constraints and the curvature value on each vertex, thereby yielding a landmark- and curvature-based surface registration in real-time.
    %\item The method could be applied for face classification with a satisfactory result.
\end{enumerate}

\section{Related Works}

\subsection{Quasi-Conformal Mapping}

% In order to formulate the distortion among different surfaces and planes, Quasi-conformal geometry is proposed by defining locally invertible and Lipschitz mappings across different complex domains~\cite{heinonen1998quasiconformal,lehto1973quasiconformal}. Benefiting from the robustness of deformations and wide domain applications, Quasi-conformal geometry is widely applied in registration~\cite{chen2021deep,guo2022automatic}, segmentation~\cite{zhang2021topology}, and shape matching~\cite{zhang2021quasi}. 

Surface parameterization and mapping problems are widely investigated, and algorithms have been proposed in recent decades~\cite{floater2005surface,sheffer2007mesh}. In particular, as conformal maps preserve the local geometry, researchers developed algorithms that could be applied to various fields~\cite{desbrun2002intrinsic,levy2002least,mullen2008spectral,yueh2017efficient}. However, the computation of conformal mapping is expensive and quasi-conformal theory has been found useful with trade-offs between conformality and other prescribed constraints~\cite{heinonen1998quasiconformal,lehto1973quasiconformal}. The Beltrami coefficient is introduced to measure the distortion of the mapping. In \cite{gardiner2000quasiconformal}, Gardiner et al. proved the 1-1 correspondence between  Beltrami coefficients and surface diffeomorphisms.

The utilization of quasi-conformal geometry in registration tasks allows for the alignment of different surfaces by minimizing distortion, which enables accurate and consistent alignment, even when dealing with complex deformations and variations in shape. In segmentation, quasi-conformal geometry has been applied to delineate boundaries and segment regions of interest by leveraging the topological properties of the underlying complex domains. Additionally, shape-matching techniques based on quasi-conformal geometry enable efficient and effective matching of shapes by preserving their local geometric properties.

By incorporating quasi-conformal geometry into our approach, we leverage its robustness and versatility to achieve accurate and reliable registration of partial face surfaces. This enables us to align and integrate partial faces with minimal distortion, facilitating subsequent analysis and recognition tasks.

\subsection{Facial Landmark Detection}

Facial landmark detection plays a crucial role in facial analysis as it provides key points representing anatomical structures, facial features, and regions of interest for various high-level tasks such as face recognition, registration, and editing~\cite{wu2019facial}. Traditional methods for landmark detection rely on machine learning algorithms, including Support Vector Machines (SVM)~\cite{uvrivcavr2012detector} and Random Forests (RF)~\cite{luo2015locating}, using manually crafted key points. However, these approaches require expert annotations and have limited generalization capabilities across different datasets.

Recent advancements in landmark detection leverage the powerful feature extraction capabilities of Convolutional Neural Networks (CNN) to perform unsupervised landmark detection~\cite{kowalski2017deep,burgos2013robust}. To further improve detection accuracy and generalization abilities, researchers have proposed improved network architectures~\cite{zou2019learning,zhao2020mobilefan}, learning patterns~\cite{zhang2014facial}, and well-designed loss functions~\cite{feng2018wing,wang2019adaptive}. However, these existing approaches still face challenges in identifying facial landmarks with significant variations or on partial faces.

In this work, we present our landmark detection network specifically designed for partial human faces with diverse variations. Our network addresses the limitations of existing methods by effectively detecting landmarks on partial faces, thereby providing crucial information for subsequent registration processes.

%Facial landmarks are the key points for facial analysis since it provides different anatomical structures, facial key points, or regions of interest for high-level analysis such as face recognition, registration, and editing~\cite{wu2019facial}. Traditional methods for landmark detection employ machine learning algorithms such as Support Vector Machine (SVM)~\cite{uvrivcavr2012detector} and Random Forest (RF)~\cite{luo2015locating} with handcrafted key points. However, these approaches require expert annotations and have limited generalization ability on different datasets. Recent methods depending on the strong feature extraction ability of Convolutional Neural Networks (CNN) conducts landmark detection in unsupervised manners~\cite{kowalski2017deep,burgos2013robust}. To further enhance detection accuracy and generalization ability, improved network architectures~\cite{zou2019learning,zhao2020mobilefan}, learning patterns~\cite{zhang2014facial}, and well-designed losses~\cite{feng2018wing,wang2019adaptive} are proposed. However, current approaches are hindered from identifying facial landmarks with huge variations or on partial faces. In this work, we present our landmark detection network for partial human faces with different variations to assist further registration. 

\subsection{Facial Registration}

% The core idea of face registration is aligning distorted facial landmarks with original ones. With L2 loss, traditional approaches for key point alignment are implemented by CNN-based networks to learn one-to-one mappings~\cite{zhang2015learning}. Then, advanced modules such as inception blocks~\cite{sun2015deepid3} and siamese networks~\cite{schroff2015facenet,sun2015deeply} are applied to extract features at different scales and learn discriminative embeddings. Furthermore, to reduce the adverse effects of huge inter- and intra- landmark loss, wing loss and triplet loss are proposed, and applied in further registration works. The other extended works include 3D surface registration~\cite{yenamandra2021i3dmm,ranjan2017hyperface} that captures distortion mapping on 3D objects by processing data with 3DMM~\cite{tran2018nonlinear}. Beyond capturing point mappings solely by neural network, Quasi-conformal methods formulate the distortion by defining distortion factor, simplifying the training objective as predicting Beltrami coefficient between domain pairs, which work for both 2D~\cite{chen2021deep} and 3D objects~\cite{guo2022automatic,zhang2021quasi}. However, both Quasi-conformal based methods and other deep learning methods struggle to align faces with partial components and non-rigid deformation. To address this issue, we propose our Quasi-conformal based registration model for free-boundary partial faces to fulfill the gaps. 

Traditional face registration methods primarily focus on aligning distorted facial landmarks with their corresponding landmarks using CNN-based networks and L2 loss \cite{zhang2015learning}. These methods often incorporate advanced techniques such as inception blocks \cite{sun2015deepid3} and siamese networks \cite{schroff2015facenet,sun2015deeply} to extract multi-scale features and learn discriminative embeddings. To handle variations in facial landmarks, techniques like wing loss and triplet loss have been proposed. 3D surface registration methods have also been developed to address non-linear distortions. These methods\cite{yenamandra2021i3dmm,tran2018nonlinear} utilize 3D models to capture complex facial deformations.

However, acquiring a complete face for registration purposes can be challenging, leading to a growing interest in methods for partial surface registration. In recent years, several approaches have been proposed to tackle this problem. In \cite{cao2023unsupervised}, Cao et al. introduced a method that utilizes deep functional maps for unsupervised training to achieve shape correspondence even for partial shapes. Bracha et al. proposed a novel idea in \cite{bracha2023partial}, where they establish a direct correspondence between partial and full face shapes through feature matching. These recent advancements in the field of partial face registration highlight the growing interest in finding effective solutions for this challenging problem. However, existing research often lacks focus on the analysis of face surfaces. Consequently, the challenges associated with feature extraction and region correspondence for partial faces remain a significant problem that requires further attention and investigation.

To address this challenge, we propose a quasi-conformal-based registration model for partial faces' correspondence. Our model combines the benefits of quasi-conformal methods while handling partial components and non-rigid deformations, filling the existing gaps in face registration research.

\section{Mathematical Background}
\subsection{Quasi-conformal theory}

In this section, we introduce the theories related to quasi-conformal maps. For more details, please refer to \cite{astala2008elliptic, gardiner2000quasiconformal}.

Compared with the conformal map, the quasi-conformal map is more generalized and can be understood to be a map with bounded conformality distortion. An orientation-preserving homeomorphism $f:\Omega \subset \mathbb{C} \rightarrow \Omega' \subset \mathbb{C}$ is said to be a quasi-conformal map if it satisfies the Beltrami equation:
\begin{align}
\label{BeltramiEqu}
\frac{\partial f}{\partial \Bar{z}} = \mu_f(z)\frac{\partial f}{\partial z},
\end{align}

where $\mu_f(z)$ is a complex-valued Lebesgue-measurable function satisfying $\|\mu_f(z)\|_\infty < 1$ is the Beltrami coefficient of $f$ and it represents the conformality distortion of $f$.

\begin{figure}[!h]
\centering
\includegraphics[width=0.7\linewidth]{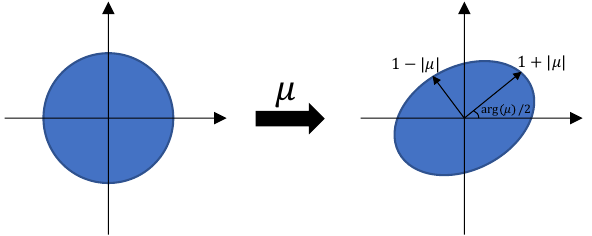}
\caption{The illustration of how Beltrami coefficient $\mu$ determine the conformal distortion.}
\label{mu}
\end{figure}

If we are given a Beltrami coefficient $\mu$ with $\|\mu\|_\infty < 1$, we can solve its corresponding quasi-conformal mapping satisfying the Beltrami equation\cite{gardiner2000quasiconformal}.

\begin{theorem}
Suppose $\mu: \mathbb{D} \rightarrow \mathbb{C}$ is Lebesgue measurable with $\| \mu \|_{\infty} < 1$. There is a quasi-conformal homeomorphism $\phi$ from $\mathbb{D}$ to itself, which is in the Sobolev space $W^{1,2}(\mathbb{D})$ and satisfies the Beltrami equation~\eqref{BeltramiEqu} in the distribution sense. Furthermore, by fixing 0 and 1, $\phi$ is uniquely determined.
\label{mu2map}
\end{theorem}

% Conversely, if the orientation preserving homeomorphism $\phi$ is presented, the corresponding Beltrami coefficient can be obtained by:
Conversely, if an orientation preserving homeomorphism $\phi$ is preseted, we can get its Beltrami coefficient by:
\begin{align}
    \mu_{\phi}=\frac{\partial \phi}{\partial \bar{z}} / \frac{\partial \phi}{\partial z}.
\end{align}

Thus, the Jacobian Matrix $J$ of $\phi$ can be expressed according to $\mu_\phi$ as follows:
\begin{equation}
J(\phi)=\left|\frac{\partial \phi}{\partial z}\right|^{2}\left(1-\left|\mu_{\phi}\right|^{2}\right).
\end{equation}

Given that $\phi$ is an orientation preserving homeomorphism, $J(\phi)>0$ and $\left|\mu_{\phi}\right|<1$ everywhere. Hence, $\left\|\mu_{\phi}\right\|_{\infty}<1$ should hold. 
Theorem \ref{mu2map} indicates that under suitable normalization, every $\mu$ with $\left\|\mu\right\|_{\infty}<1$ is associated with a unique homeomorphism. Therefore, a homeomorphism from $\mathbb{C}$ or $\mathbb{D}$ onto itself can be uniquely determined by its associated Beltrami coefficient.

In addition, we can also composite two quasi-conformal maps and the Beltrami coefficient can be computed. Suppose we have $f, g$: $\mathbb{C} \rightarrow \mathbb{C}$, two quasi-conformal maps, associated with Beltrami coefficients $\mu_f$ and $\mu_g$ respectively. Thus, the Beltrami coefficient of $g \circ f$ is
\begin{equation}
\mu_{g\circ f} = \frac{\mu_f + (\mu_g \circ f)\tau}{1 + \Bar{\mu}_f(\mu_g\circ f)\tau}, \tau = \frac{\Bar{f}_z}{f_z}
\end{equation}

\subsection{Conformal Parameterization}

In this section, we are going to introduce the variational formulation of the conformal parameterization from \cite{desbrun2002intrinsic}. 

Suppose we have a piecewise linear mesh patch $M$ and we want to construct a piecewise linear mapping $\psi$ between $M$ and a planar triangulation $U\in \mathcal{R}$ conformally. We denote $x_i$ to be 3D positions of the i-th vertex on $M$ and $u_i$ is its corresponding position on $U$. 

In \cite{desbrun2002intrinsic}, they introduced the Dirichlet energy and Chi energy on triangulations.
\begin{align}
    E_A = \sum_{oriented\,edges(i, j)} \cot \alpha_{ij}|u_i - u_j|^2,\\
    E_\chi = \sum_{j\in N(i)} \frac{\cot \gamma_{ij} + \cot\delta_{ij}}{|x_i - x_j|^2}(u_i - u_j)^2
\end{align}
in which $\alpha_{ij}$ is the opposite left angle of edge $(i,j)$ on the same triangular, $\gamma_{ij}$ and $\delta_{ij}$ are the angles on the both side of edge $(i,j)$ near vertex $x_j$. In this way, the distortion measure $E$ can be defined to be 
\begin{align}
    E = \lambda E_A + \zeta E_\chi
\end{align}
in which $\lambda$ and $\zeta$ are parameters. By minimizing $E$, we can get a smooth parameterization.

\subsection{Surface Curvature}

Curvature is an important quantity in differential geometry for assessing how a surface deviates from a plane. In the human face surface, the curvature values contain the personal features that could be used for recognition.

We define $N : S \rightarrow \mathbb{S}^2 \subseteq \mathbb{R}^3$ to be the \emph{normal map} giving unit vector at each point $p$. Suppose $C$ is a regular curve on $S$, $p$ is a point on $S$ and $k$ is the curvature of $C$ at $p$. We set $\cos \theta = \langle n, N \rangle$, where $N$ is the normal vector to $S$ at $p$ and $n$ is normal to $C$. $k_n = k \cos \theta$ is called the \emph{normal curvature} of $C$ at $p$. The \emph{principal curvatures} at $p$ are the maximum and minimum of the normal curvature, denoted as $k_1$ and $k_2$ respectively. The \emph{mean curvature} at $p$ is defined to be $H = \frac{1}{2}(k_1 + k_2)$.

\begin{figure}[!h]
\centering
\includegraphics[width=0.5\linewidth]{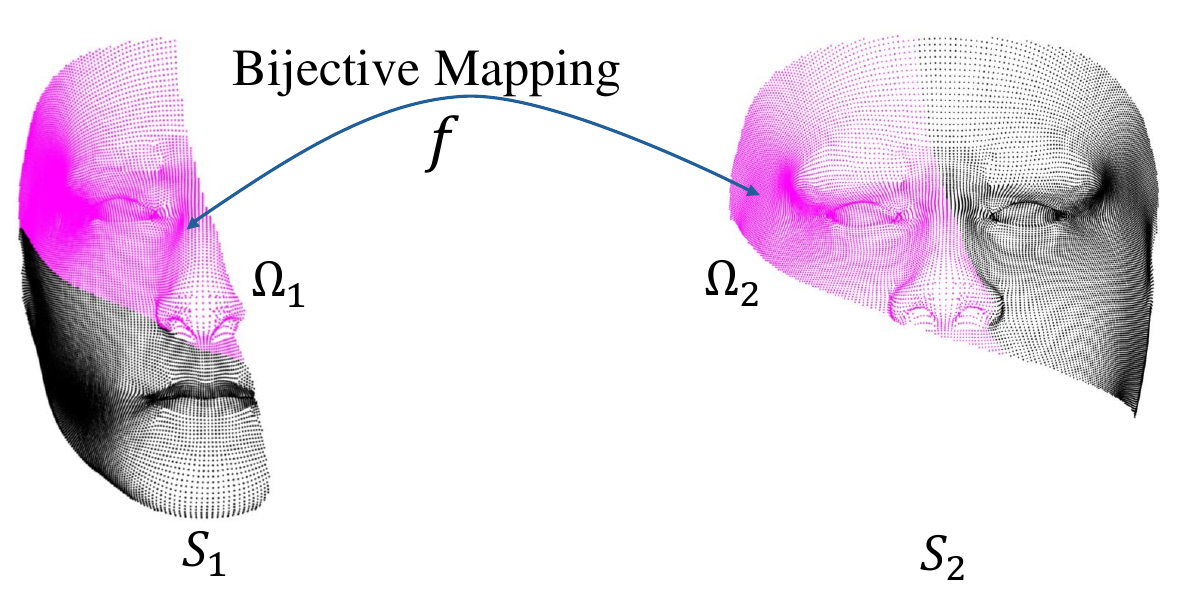}
\caption{Demonstration of the surface matching problem. $S_1$ and $S_2$ are two partial faces with similar parts. $\Omega_1$ and $\Omega_2$ are the intercepted parts and labeled in magenta. The goal is to find a bijective mapping $f$ such that the regional correspondence can be constructed for $\Omega_1$ and $\Omega_2$.}
\label{demo_f}
\end{figure}

\section{Deep Surface Registration of Partial Faces}

The problem of partial surface correspondence and registration is highly significant, particularly in scenarios with limited data availability, as not all situations are ideal. To tackle this problem, we initiate our analysis by introducing the subsequent matching problem of partial surfaces.

\begin{problem}
    Given two surfaces $S_1$ and $S_2$, where $S_1$ and $S_2$ has similar regions. Suppose there exist regions $\Omega_1$ and $\Omega_2$ such that $\Omega_1 \subseteq S_1$ and $\Omega_2 \subseteq S_2$ respectively with similar features, the problem is to find an optimal bijective mapping $f: \Omega_1 \rightarrow \Omega_2$ such that $f(\Omega_1) = \Omega_2$.
\end{problem}

To visualize this problem, we demonstrate it using human faces as examples in Fig. \ref{demo_f}. In this paper, we are introducing a deep learning based method for finding the optimal bijective mapping $f$ for partial faces' matching problem, which is then applied to facial recognition.

\subsection{Framework Overview}

\begin{figure}[!h]
\centering
\includegraphics[width=0.5\linewidth]{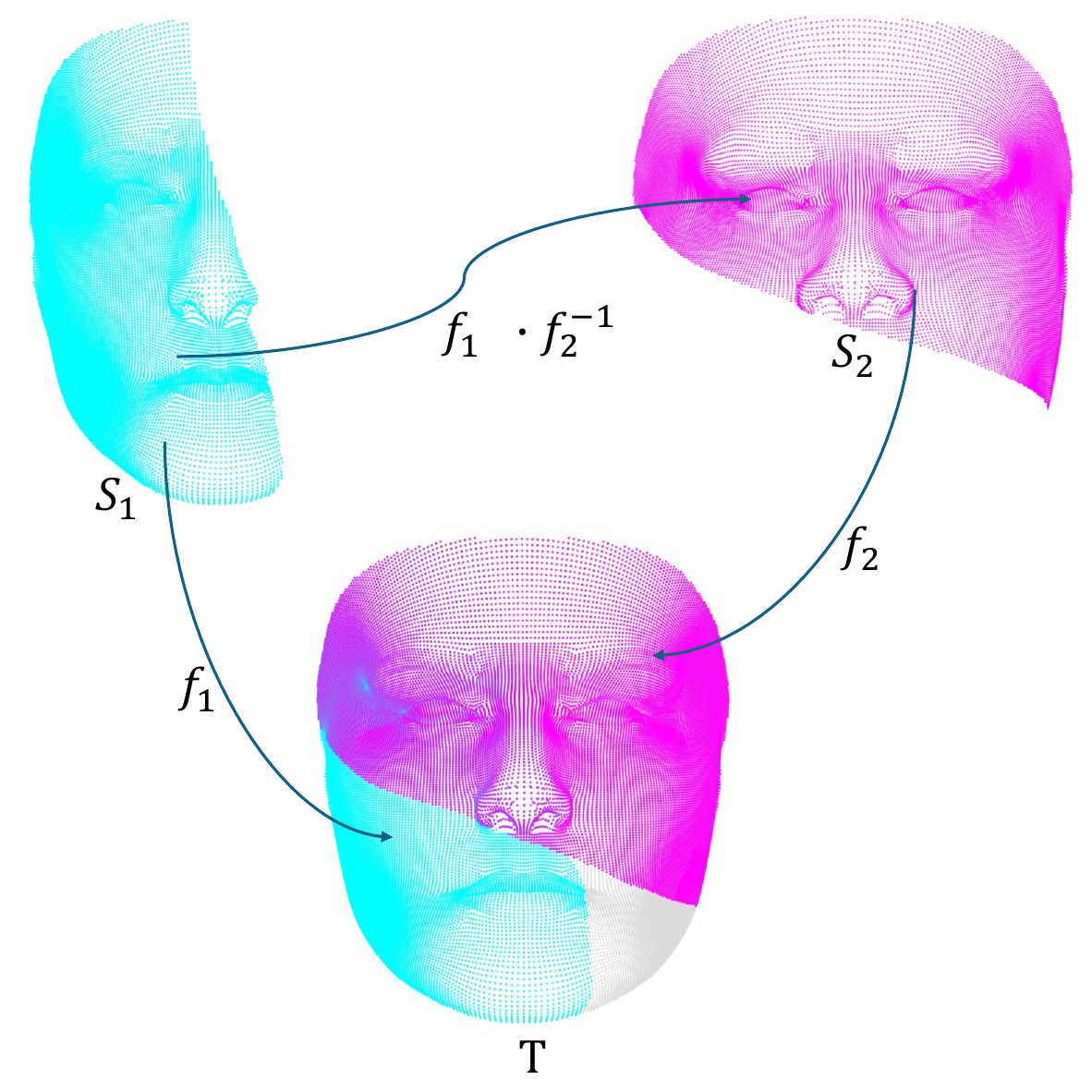}
\caption{Demonstration of the strategy for solving partial surface correspondence problem. $S_1$ and $S_2$ are two partial faces in cyan and magenta, respectively. $T$ is the template face in gray. $f_1$ and $f_2$ are bijective mappings.}
\label{overview}
\end{figure}

In this section, we present a detailed overview of our framework for partial face registration, and the strategy is shown in Fig. \ref{overview}. 

Our method aims to take a mean face as the template and find the optimal mapping that registers the partial face to the template. In Fig. \ref{overview}, $S_1$ and $S_2$ are two partial faces that share similar regions. For each partial face, our model could generate a quasi-conformal mapping $f_1$, which is bijective, and map the partial face $S_1$ to the corresponding region on template face $T$. By applying the same method to the partial face $S_2$, we obtain a bijective quasi-conformal mapping $f_2$. Consequently, we can compute an inverse mapping of $f_2$ and compose it with $f_1$. This enables us to register the partial face $S_1$ to $S_2$ by mapping $f_1 \cdot f_2^{-1}$.

%Our approach involves utilizing a mean face as the template to facilitate the registration process. Additionally, we leverage neural networks to enable real-time processing capabilities. 

\subsection{Method Overview}

\begin{figure}
\centering
\includegraphics[width=1\linewidth]{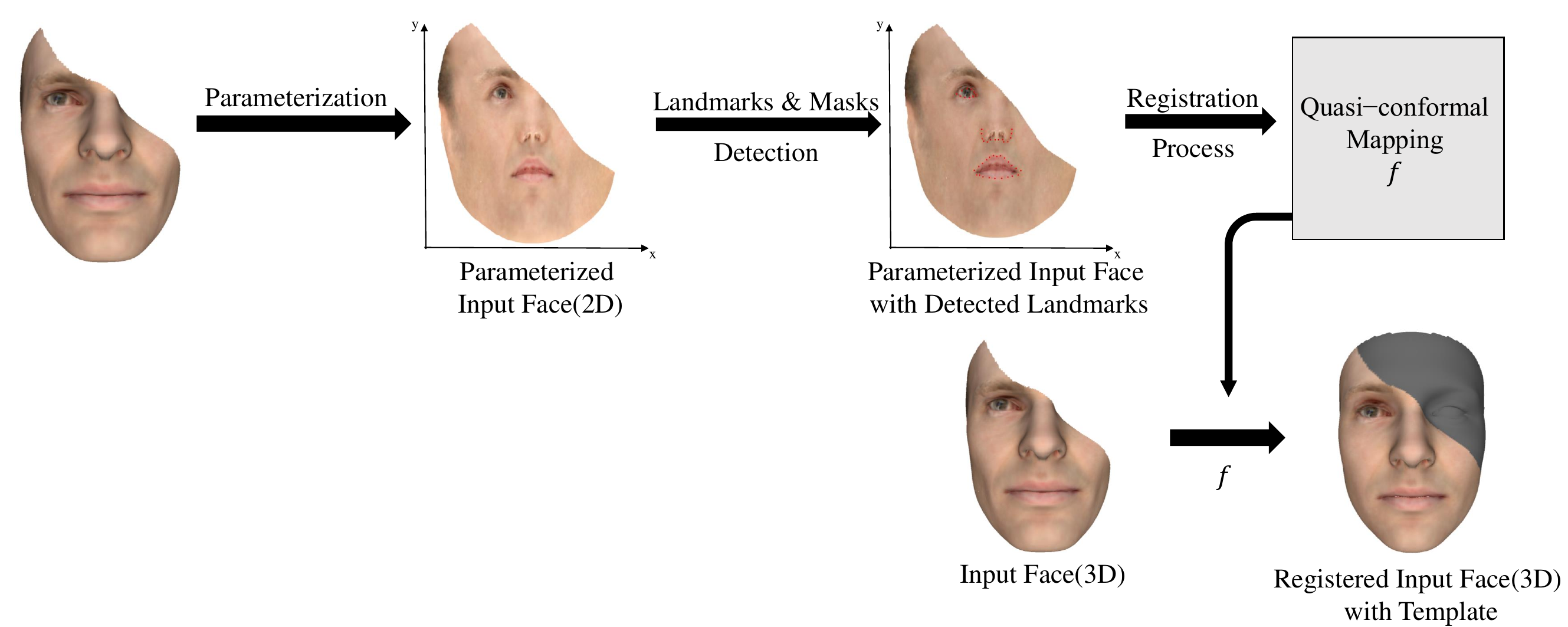}
\caption{Demonstration of the framework for registration. The first row demonstrates how we generate a quasi-conformal mapping based on the input face mesh. Then, we demonstrate how the mapping registers an input face mesh to the template face.}
\label{Whole}
\end{figure}

In this section, we introduce the details of finding the optimal mapping $f$ for mapping a partial face to the corresponding region on the template face $T$. The overview of it is shown in Fig. \ref{Whole}. Initially, we parameterize the partial face to the 2D domain by conformal and quasi-conformal mapping with the least angle distortion. Subsequently, a network is constructed to identify the landmarks of the parameterized face and ascertain the existence of each point. The final step involves feeding the output from the preceding network, along with the parameterized partial face, into the registration network. This process facilitates the acquisition of appropriate Beltrami Coefficients, which in turn generate a corresponding quasi-conformal mapping to register the input face to the template face, as shown in Fig. \ref{Whole}. Details for each component will be illustrated in the following parts.

% In this section, we introduce the details of our framework. Firstly, we parameterize the 3D partial face mesh on the 2D domain by using quasi-conformal theory to preserve the least distortion of the mesh. Then we build a network to detect the landmarks of the parameterized face and determine each point's existence. Finally, we take the output from the previous network together with the parameterized partial face as input for the registration network, and we could get suitable Beltrami Coefficients for generating a corresponding quasi-conformal mapping. The whole framework could be seen in Fig. \ref{Whole}.

%This section provides an in-depth overview of our framework. The aim of our work is to register different faces to the template face and then conduct recognition between them. To achieve this, we separate it into three steps.  By registering different partial faces to the template, we can obtain the correspondence between them. Then, we could apply this method to different faces separately and could perform recognition as shown in Fig. \ref{reg}.

\subsubsection{Parameterization Method}

\begin{figure}[!h]
\centering
\includegraphics[width=1\linewidth]{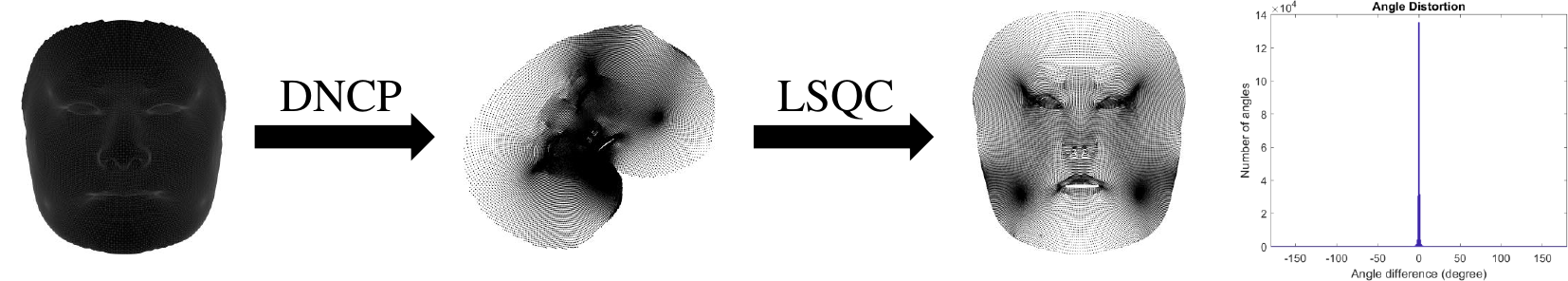}
\caption{The flow on the left demonstrates the parameterization by discrete natural conformal parameterization(DNCP) and modified by least-squares quasi-conformal map (LSQC). The chart on the right shows the angle distortion which is small.}
\label{Para}
\end{figure}

The partial face is a 3D surface mesh, and to adopt the learned method, we need to parameterize it into the 2D domain. We parameterize the 3D mesh conformally and then use the quasi-conformal theory to get a visualizable face mesh while minimizing the angle distortion and keeping the majority of information of the surface mesh as shown in Fig. \ref{Para}. 

Initially, we adopt the Discrete Natural Conformal Parameterization(DNCP) in \cite{desbrun2002intrinsic} and thus get a free boundary map from the surface. The mapping generated by DNCP is conformal, which preserves every angle of the mesh. However, as shown in Fig. \ref{Para}, the parameterized face is hard to visualize as a face and also hard to determine the location of eyes or mouth; thus, we are going to use quasi-conformal theory to modify it by using an approach called least-squares quasi-conformal map (LSQC)\cite{qiu2019computing, zhu2022parallelizable} with proper boundary conditions. Suppose $f:\Omega\subset\mathbb{C} \rightarrow \Omega'\subset\mathbb{C}$ is a quasi-conformal map and $f = u+iv$, $\mu_f = \rho + i\tau$ where $u, v, \rho$ and $\tau$ are real-valued functions. We can transfer the Beltrami equation to
\begin{align}
    \begin{pmatrix}u_x \\ u_y\end{pmatrix} = \begin{pmatrix}0 & 1 \\ -1 & 0 \end{pmatrix} A \begin{pmatrix}v_x \\ v_y\end{pmatrix},
\end{align}
in which 
\begin{align}
    A = \frac{1}{1 - |\mu|^2}\begin{pmatrix} (\rho - 1)^2+\tau^2 & -2\tau \\\ -2\tau & (1+\rho)^2 + \tau^2 \end{pmatrix}.
\end{align}

By the relation $u_{xy} = u_{yx}$, we can get the equation 
\begin{align}
    \nabla\cdot(A\nabla v(z)) = 0, \, \nabla\cdot(A\nabla u(z)) = 0.
\end{align}

These are the Euler-Lagrange equations of the following two Dirichelet-type energies respectively:
\begin{align}
    E_A(u) = \frac{1}{2}\int_\Omega\|A^{1/2}\nabla u\|^2, \, E_A(v) = \frac{1}{2}\int_\Omega\|A^{1/2}\nabla u\|^2.
\end{align}

Here we introduce the energy $E_{LSQC}$ from \cite{qiu2019computing}, and we set  the boundary of $\Omega$ to be mapped to a given boundary and get a quasi-conformal map from it by minimizing:
\begin{align}\left\{\begin{aligned}
    & E_{LSQC}(u, v) = \frac{1}{2}\int_\Omega \|P\nabla u + JP\nabla v\|^2dxdy \\
    & f(\partial \Omega) = \partial \Omega'
\end{aligned}\right. ,\end{align}
in which 
\begin{align}
    P = \frac{1}{\sqrt{1 - |\mu|^2}}\begin{pmatrix}
        1- \rho & -\tau \\ -\tau & 1 + \rho
    \end{pmatrix} \,\, \text{and} \,\, J = \begin{pmatrix}
        0 & -1 \\ 1 & 0
    \end{pmatrix}
\end{align}

It is observed that $P^TP = A$ and the Beltrami equation holds if and only if $E_{LSQC}(u, v) = 0$. There is a relationship between $E_A(u)$, $E_A(v)$, $E_{LSQC}(u, v)$ and $\mathcal{A}(u, v)$. $\mathcal{A}(u, v)$ is the area of $\Omega' = f(\Omega)$:
\begin{align}
E_A(u) + E_A(v) - E_{LSQC}(u, v) = \mathcal{A}(u, v) = \int_\Omega(u_xv_y - v_xu_y)dxdy
\label{AreaEqu}
\end{align}

In our setting, the face surface is discretized, and $f$ is linear on each triangular face. Thus, we have $\mu$ constant on each face. On an oriented triangular face $T=[w_0, w_1, w_2]$, the gradient of a function $f=(f_{0}, f_{1}, f_{2})$ is given by:
\begin{align}
    \nabla f = \frac{1}{2Area(T)}\begin{pmatrix}
        0 & -1 \\ 1 & 0 
    \end{pmatrix} \sum_{i=0, 1, 2}f_{i}(w_{2+i} - w_{1+i})
\end{align}
where the index should modulo 3. In this way, $E_A(u)$ and $E_A(v)$ can be discretized by summing over all faces which could be expressed with the generalized Laplacian matrix $\mathcal{L}_\mu$:
\begin{align}
    E_A(u) = u^T\mathcal{L}_\mu u,\, E_A(v) = v^T\mathcal{L}_\mu v.
\end{align}

We can also discretize the area matrix from Equ.\ref{AreaEqu} by some skew-symmetric matrix $U$:
\begin{align}
    \mathcal{A}(u, v) = \begin{pmatrix} u^T & v^T \end{pmatrix} \begin{pmatrix} 0 & U \\ -U & 0 \end{pmatrix} \begin{pmatrix} u \\ v \end{pmatrix}
\end{align}

We can derive a symmetric matrix $N$:
\begin{align}
    N = \begin{pmatrix} \mathcal{L}_\mu & 0 \\ 0 & \mathcal{L}_\mu \end{pmatrix} 
 - \begin{pmatrix} 0 & U \\ -U & 0 \end{pmatrix}
\end{align}

In this way, the energy $E_{LSQC}$ can be discretized to be:
\begin{align}
    E_{LSQC}(u,v) =\begin{pmatrix} u^T & v^T \end{pmatrix} N \begin{pmatrix} u \\ v \end{pmatrix}
\end{align}

To solve the corresponding mapping, we set $\mu=0$ with landmark constraints. The aim of this step is to have a 2D face mesh with a proper shape for visualization. Suppose the input 3D face mesh is $M_0$ and the boundary points are $\partial M_0 = \{m_0, m_1, ..., m_k\}$, in which $m_i = (x_i, y_i, z_i)$. We project the boundary points of the face mesh to the 2D domain as $\hat{m}_i = (x_i, y_i)$. Since we parameterize the mesh $M_0$ by DNCP first, we denote the parameterized face mesh to be $M_1$ and the boundary points are $\partial M_1 = \{n_0, n_1, ..., n_k\}$. In this way, we are solving the following problem:
\begin{align}\left\{\begin{aligned}
    &E_{LSQC}(u, v) = \begin{pmatrix} u^T & v^T \end{pmatrix} N \begin{pmatrix} u \\ v \end{pmatrix} \\
    &f(n_i) = \hat{m}_i, \,\,\,\,\,n_i \in \partial M_1
\end{aligned}\right. .\end{align}

Minimizing the energy $E_{LSQC}$ could generate a mapping satisfying the input value $\mu$ if there are no boundary constraints. However, due to the added boundary conditions, very few angle distortions are introduced. In this way, the Beltrami coefficients $\mu$ would close to zero, and the mapping could be quasi-conformal, as shown in Fig. \ref{Para}.

%The relation could also be written as:

%\begin{align}
%    E_A(u) + E_A(v) - E_{LSQC}(u, v) = \mathcal{A}(u, v) = \int_\Omega(u_xv_y - v_xu_y)dxdy.
%\end{align}

\subsubsection{Landmark Detection}

%In this section, we describe our proposed architecture for Landmark Detection Network (LD-net) for detecting prominent points on the parameterized partial face.

%In the previous section, we have already introduced the pipeline for flattening the partial face by quasi-conformal theory. For each vertex, we also compute its curvature and store it for further usage. To adopt the capability of the convolutional neural networks, we change the parameterized data into the image domain, and the value of the pixel will be determined by its curvature value. After this process, we can easily take it as input for our designed network for landmark detection.

%Since the input data is the partial face, the goal for this task is to determine not only locations but also the existence of each point. The structure of our proposed network is shown in Fig. \ref{LDnet}.

In this section, we outline our proposed Landmark Detection Network (LD-net) architecture, designed specifically for detecting significant points on a parameterized partial face. The detected landmarks will be used to assist the registration process.

We previously introduced the process for flattening the partial face using quasi-conformal theory. We also computed the curvature for each vertex, storing this data for future use. To leverage the capabilities of convolutional neural networks, we transformed the parameterized data into the image domain, with pixel values determined by their corresponding curvature values. The rest of the pixels are remaining zero. This transformation allows us to easily use the data as input for our landmark detection network.

\begin{figure}[!h]
\centering
\includegraphics[width=0.8\linewidth]{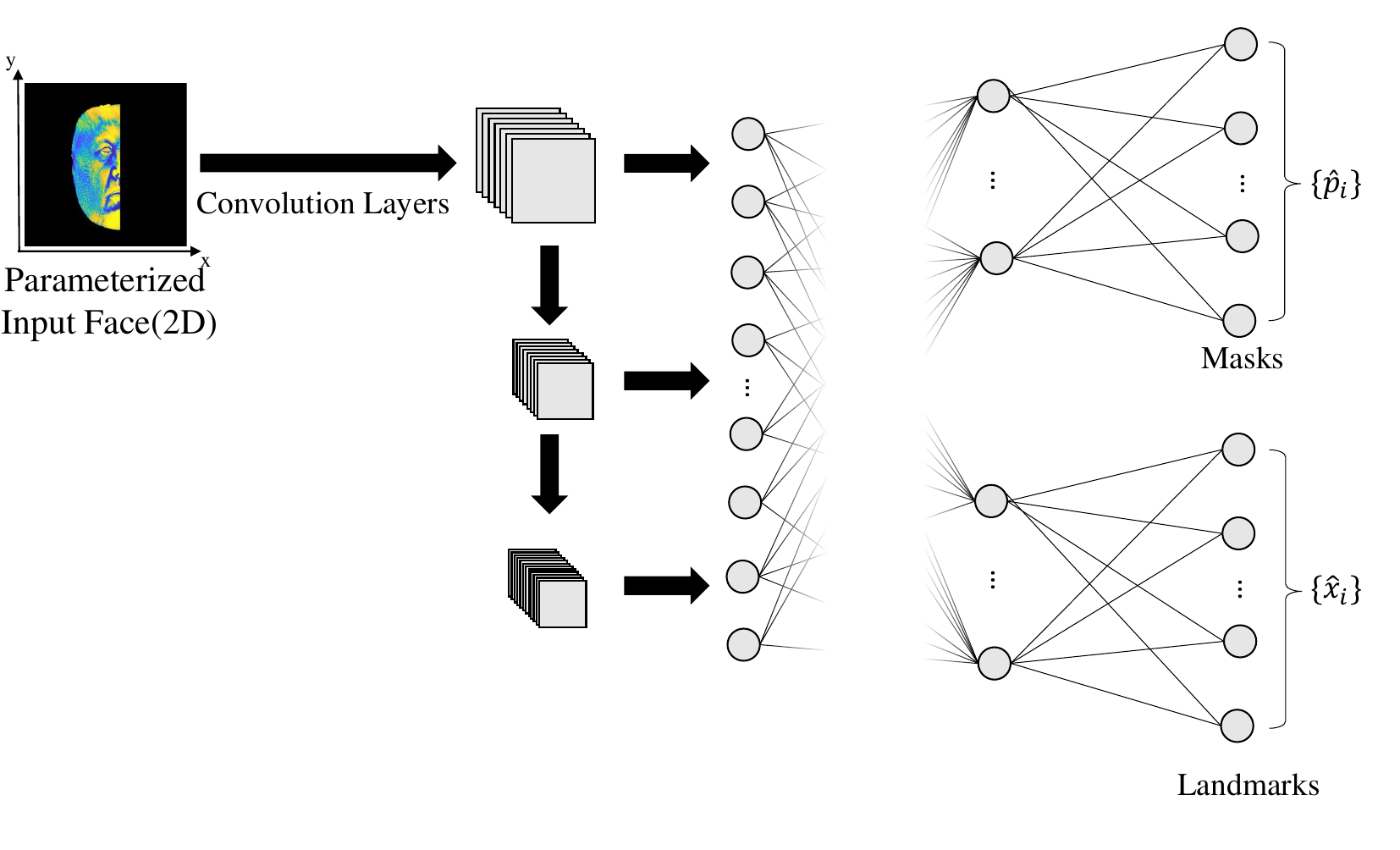}
\caption{Demonstration for the LD-net. The input parameterized partial faces pass through convolution layers, and we take features from different scales and concatenate them into a column. Then we take the column data into two different multilayer perceptrons then generate the output masks and landmark locations. } 
\label{LDnet}
\end{figure}

Besides recording the vertex value on the image domain, we also save the location and the existence of the landmark points as labels for the supervised training. In the inference part, the input data is a partial face on the image domain, and the objective of this network extends beyond merely determining point locations; it also involves confirming the presence of each point. The structure of our proposed network is illustrated in Fig. \ref{LDnet}.

We first pass the smoothed image with curvature value on each vertex into convolution layers as shown in Fig. \ref{LDnet}. Inspired by the structure by \cite{guo2019pfld}, we take the mean value of each channel at different stages and concatenate it to be a column for further process. The advantage is that the features at different scales are all taken into the multi-layer perceptron(MLP). Next, we take the concatenated value to two MLPs separately, and MLPs could learn from the features to generate the location and the existence of the landmarks. In the inference part, we take the detected prominent points with probability over 90\% as existing.

\textbf{Loss Function:} For the training of LD-net, we adopt a supervised mode. The loss function of this network is as follows:
\begin{equation}
\label{LD_loss}
\mathcal{L}_{\text{LD}} =  \frac{\alpha}{N}\sum_i^N -[\hat{p}_i \cdot \log(p_i) + (1 - \hat{p}_i) \cdot \log(1 - p_i)] +  \frac{\beta}{N}\sum_i^N \|(\hat{x}_i - x_i) \odot p_i\|,
\end{equation}

in which $N$ is the number of prominent points predefined on the face, $\hat{p}_i$ and $\hat{x}_i$ are the output mask probability and landmark locations for the i-th point from the network, $p_i$ and $x_i$ are the label mask and label location for the i-th point. In our setting, the predicted $\hat{p}_i \in [0, 1]$ and the labeled $p_i = 0$ or $1$.

The first term in Equ. \ref{LD_loss} is designed for the existence prediction. The network generated the probability of the existence of each point, and we adopted the cross-entropy loss between the output and the label in the training part. The second term constrains the distance between the output location and the labeled locations.

\subsubsection{Registration}

 As introduced in the Section Mathematical Background, a quasi-conformal mapping is bijective when the Beltrami coefficient $\mu$ is less than 1, and we can easily get its inverse mapping. Based on this theory, we will introduce the procedure for registration between different partial face meshes in this part.

\begin{figure}[!h]
\centering
\includegraphics[width=\linewidth]{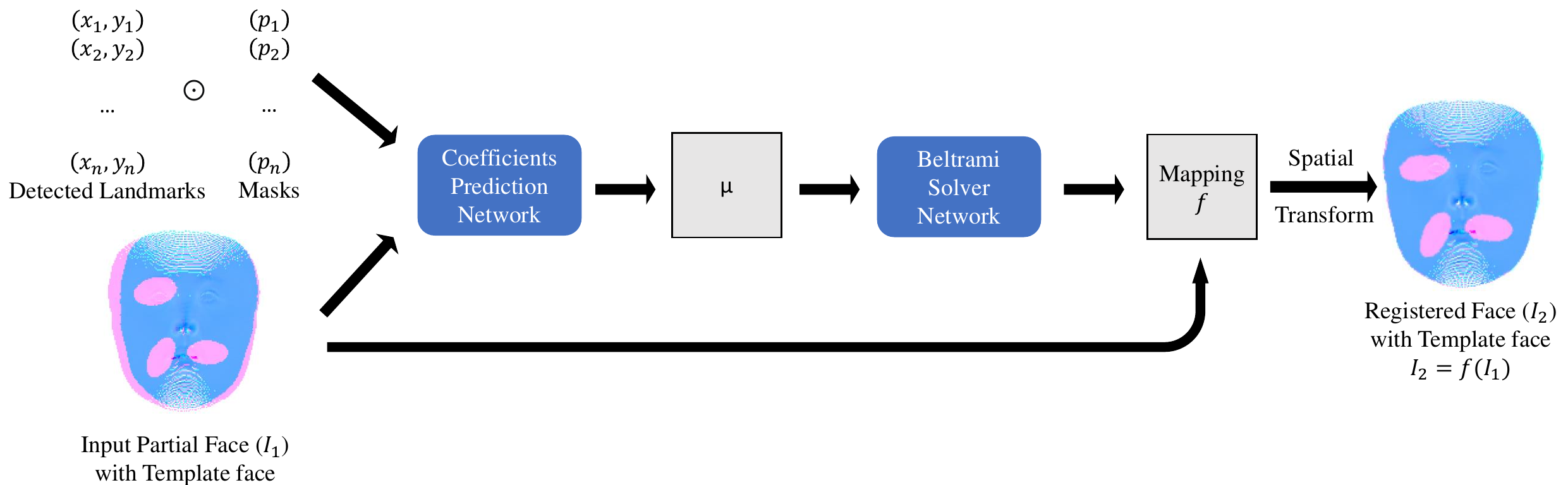}
\caption{Demonstration for the registration process. One path takes the detected landmarks' locations and masks as input, and the second path's input is the partial face. Template face is NOT the input of the Coefficient Prediction Network(CP-net). Here is for visualization and comparison. The CP-net generates the Beltrami Coefficient $\mu$, and the Beltrami Solver Network generates the quasi-conformal mapping $f$ from $\mu$. Then we can register the input partial face by $f$ to the template face.} 
\label{Reg_process}
\end{figure}

\begin{figure}[!h]
\centering
\includegraphics[width=\linewidth]{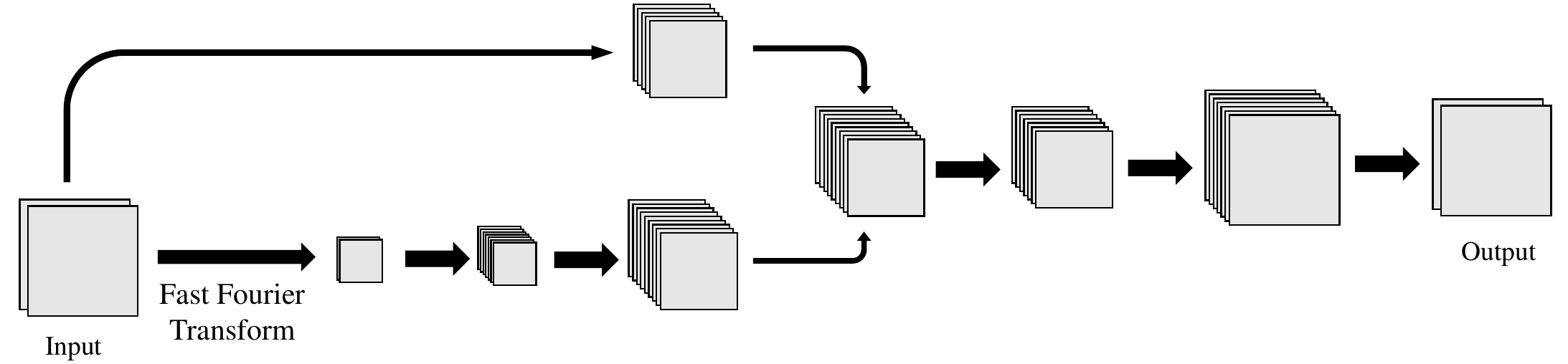}
\caption{Demonstration of the structure of BSnet from \cite{chen2021deep}.} 
\label{BSnet}
\end{figure}

 Initially, as shown in Fig. \ref{Reg_process}, we define the mean face as the template. Then, we design a network that is able to take the parameterized face and detected landmarks as input, and by using the Coefficients Prediction Network, we generate the Beltrami coefficients for producing the quasi-conformal mapping to register the input face to the template. To generate the quasi-conformal mapping in a fast and accurate way, we use the Beltrami Solver Network introduced by Chen et al.\cite{chen2021deep}. We demonstrate its structure in Fig. \ref{BSnet}. In this way, different face meshes could be registered to the template mesh, and thus, we can easily compute its inverse mapping, which allows us to compare differences between face meshes.

\begin{figure}[!h]
\centering
\includegraphics[width=0.95\linewidth]{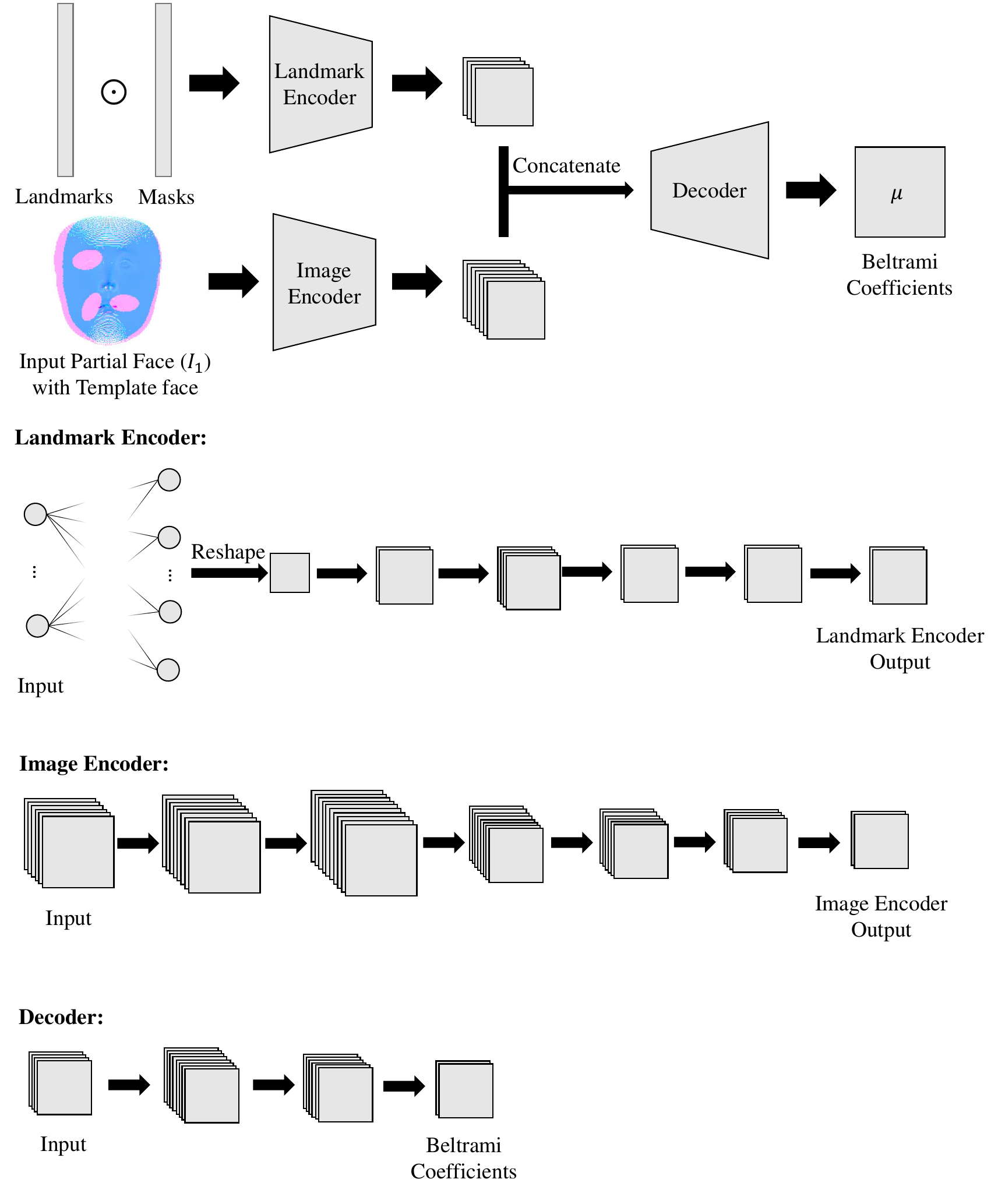}
\caption{Demonstration for the Coefficient Prediction Network. The input is separated into two paths. The first one is taking landmarks and masks as input to the Landmark Encoder, which consists of the fully connected layers, and then reshaping the output to 2D. The second path takes the input partial face into the Image Encoder which consists of the convolutional neural networks. Then, the reshaped features are concatenated and passed through the Decoder to get the Beltrami coefficients $\mu$. The Decoder consists of the transposed convolutions, and we use an activation function to keep each output value lower than 1 to generate a bijective quasi-conformal mapping.}
\label{CPnet}
\end{figure}

 Next, we introduce the details of the Coefficient Prediction Network as shown in Fig. \ref{CPnet}.  We separate the input into two paths for landmarks and parameterized faces. For the Landmark Encoder, we take the detected landmarks and masks as input and use fully connected layers to process and then reshape them into 2D channels. The input partial face on the second path is in the image domain, and we pass it through the convolutional layers and concatenate it with the reshaped output from the Landmark Encoder. To preserve the bijectivity and keep the least distortion in the generated mapping, we use an activation function at the end of the Decoder to suppress the norm of Beltrami Coefficients to be less than 1 which represents no folding, and the lower, the better.

The following activation function $\Phi$ is used:

$$
\Phi(m)(T) = \frac{e^{|m(T)|} - e^{-|m(T)|}}{e^{|m(T)|} + e^{-|m(T)|}}e^{iarg(m(T))},
$$
in which $m$ is the output from the last layer and $m(T)$ is a complex number on triangular face $T$. In this way, We have $\|\Phi(m)\|_\infty < 1$.

\textbf{Loss Function:} To solve the quasi-conformal mapping from the generated Beltrami Coefficent $\mu$, we use the BS-Net from \cite{chen2021deep} which is pre-trained. In the training process, we take $\mu$, $\nabla\mu$, landmark loss, and curvature loss into consideration, which is specified in Equ. \ref{Reg_loss}. The usage of $\mu$ and $\nabla\mu$ in the first two terms of Equ. \ref{Reg_loss} is to minimize the distortion and have a smooth map for registration. The third term is the landmark loss and the aim is to register the existing prominent points to their corresponding target. The last term is the curvature loss, which ensures the partial face can be registered even with no 1-to-1 correspondence. Thus, in the computation of the last term, we only compute the intercepted parts of the registered face and the template face.
\begin{equation}
\label{Reg_loss}
    \mathcal{L}_{Reg} =\frac{\kappa}{N}\sum^{N}_{i=1}\|\mu_i\|^2_2 + \frac{\tau}{N}\sum^{N}_{i=1}\|\nabla\mu_i\|^2_2 +  \frac{\varsigma}{M}\sum^{M}_{i=1}\|(x_i - \hat{x}_i)\odot p_i\|^2_2 + \sigma(I_2\cdot P_{I_2 \cap (I_1\cdot f^{\mu})} - I_1 \cdot f^{\mu})^2
\end{equation}
in which $x_i$ is the target points and $\hat{x}_i$ are the registrated points. $M$ and $N$ are the number of prominent points and the number of triangular faces separately. $p_i$ is the input mask for the prominent point and $p_i = 0$ or $1$. If the output probability from the LD-net is larger than 90\%, $p_i =1$. Otherwise $p_i = 0$.  Since the  $\mu_i$ and $\nabla\mu_i$ are defined on each triangular face. $I_1$ represents the starting face's curvature, and $I_2$ represents the channel of the target face's curvature. $P_{I_2 \cap (I_1\cdot f^{\mu})}$ is the mask for overlapping points of the target face mesh $I_2$ with the registered face $I_1\cdot f^{\mu}$. $f^{\mu}$ is the mapping generated from $\mu$ by BSnet.

\section{Recognition}

\begin{figure}[!h]
\centering
\includegraphics[width=\linewidth]{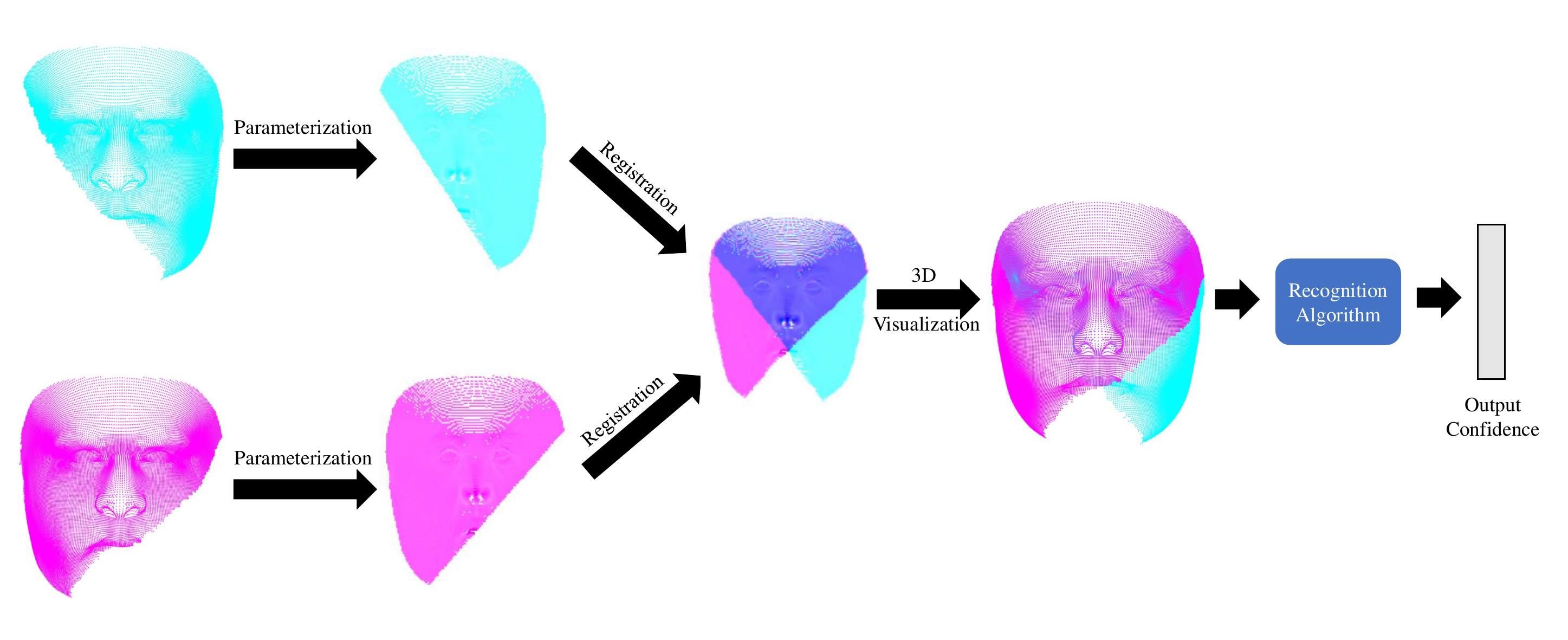}
\caption{Demonstration for the registration process of two faces. The first column consists of two input 3D faces and the second column are their corresponding parameterized faces. The third column is the registered result after two faces pass through our model separately. The fourth column is the 3D visualization of the registered faces. We use the intercepted landmarks' normal vector to compute the distance and generate the recognition confidence. }
\label{reg}
\end{figure}

With our approach, we are able to determine the existence and location of features in each input partial face, facilitating their registration to the template. By aligning the registered partial faces, we can perform recognition through the overlapping of features. We demonstrate it in Fig. \ref{reg}.

To perform the 3D face recognition, we adopt the algorithm proposed by Marras et al.\cite{marras2012robust} which introduced how to recognize a face by using the azimuth angle of surface normals. In this part, we use the facial features for recognition. The normal field is defined as a set of local surface normals $n(\boldsymbol{x}) = (n_x(\boldsymbol{x}), n_y(\boldsymbol{x}), n_z(\boldsymbol{x}))$ and $\|n(\boldsymbol{x})\| = 1$, where $\|\cdot\|$ is the $\mathcal{L}_2$ norm. The azimuth angle is defined as $\phi(\boldsymbol{x}) = \arctan \frac{n_y(\boldsymbol{x})}{n_x(\boldsymbol{x})}$. Then we can define the cosine-based dissimilarity measure between two vectors of azimuth angles $\boldsymbol{\phi}_i$ and $\boldsymbol{\phi}_j$ as:
$$d^2(\boldsymbol{\phi}_i, \boldsymbol{\phi}_j) \triangleq \sum_{\boldsymbol{x}} \{1 - cos[\phi_i(\boldsymbol{x}) - \phi_j(\boldsymbol{x})]\} $$
in which $\boldsymbol{x}$ are selected landmarks for recognition. Details for the experiment will be shown in the following part.

\section{Experiments}

\subsection{Landmark Detection}
\begin{figure}[t!]
\centering
\includegraphics[width=0.7\linewidth]{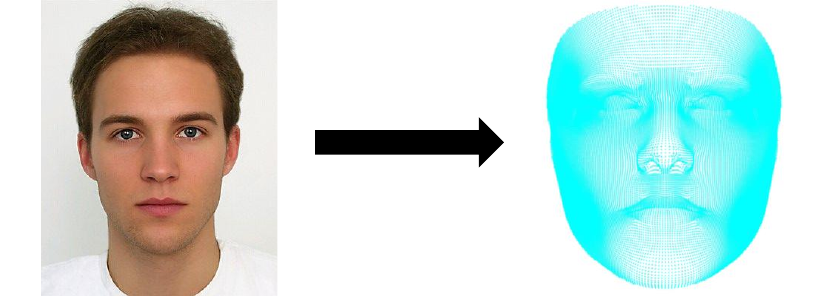}
\caption{Demonstration for generating 3D meshes from image \cite{deng2019accurate}. The image on the left is the original one and the image on the right is the mesh generated from the input image.}
\label{3dmm}
\end{figure}

\textbf{Dataset.} To train our model, we first generate faces by an CelebA dataset~\cite{liu2015faceattributes} and 3D Morphable Model (3DMM)\cite{deng2019accurate}. To start, we generate the reconstructed faces from the image as shown in \ref{3dmm}.

\begin{figure}
\centering
\includegraphics[width=\linewidth]{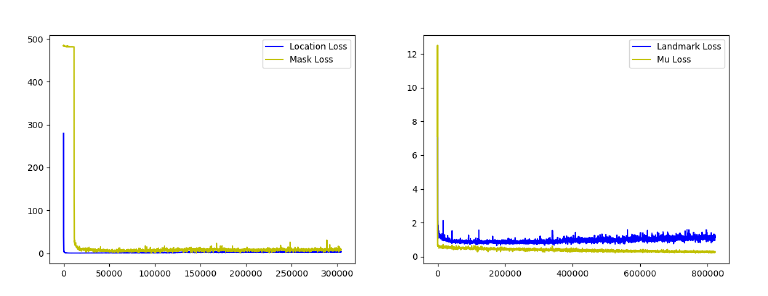}
\caption{Demonstration for the training loss of the LD-net(Left) and Reg-net(Right). The value on the vertical axis is the error and the value on the horizontal axis is the training step.}
\label{loss}
\end{figure}

We generate 20000 face meshes and randomly cut the faces in different ways. The first dataset is generated by cutting off part of the faces, which means the output is still genus-1. The second dataset is generated by cutting off holes in random sizes and locations on the face. Then, the generated parameterized surfaces are interpolated on the image domain with its mean curvature value on each vertex.

\textbf{Benchmarks.} We evaluate our model on the above-mentioned generated datasets. The interpolated images are the size of 256 by 256. The mesh surface template consists of 35709 vertexes and 70865 facets. To increase the robustness of our model, every parameterized face will be rotated and translated randomly. The training data and testing data are in the ratio of 5:1.

In training, we train the location-predicting path for the initial 50 epochs by using only landmark constraints. Then we add the mask loss gradually from epoch 50. Here you can see the loss in Fig. \ref{loss}.

\begin{figure}[t!]
\centering
\includegraphics[width=0.9\linewidth]{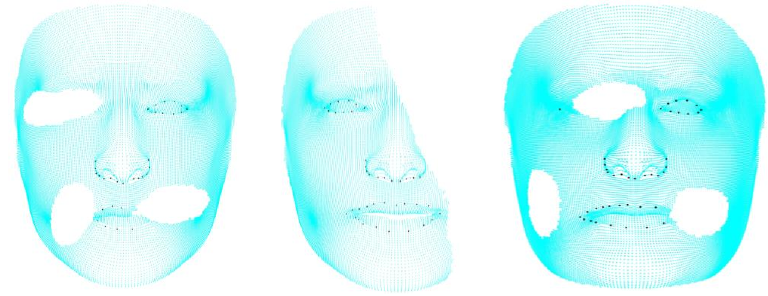}
\caption{Demonstration for landmarks detection. The black points are the detected facial features.}
\label{detect}
\end{figure}

\begin{table}[]
    \centering
    \resizebox{0.9\textwidth}{22mm}{\begin{tabular}{c|c|c|c|c|c}
        &Method & Landmark Loss & Std. of Landmark Loss & Mask Accuracy & Std. of Mask  \\
        \hline
        \multirow{5}*{Partial} &LD-net & \textcolor{red}{0.935} & 0.057 & \textcolor{red}{98.2\%} & 0.46\% \\
        %\cline{}
        ~ &SBR \cite{dong2018sbr} & 87.537 & 0.025 & $\backslash$& $\backslash$\\
        %\cline{}
        ~ &AWing \cite{wang2019adaptive} & 13.354 & 0.151 &$\backslash$ &$\backslash$ \\
        %\cline{}
        ~ &PFLD \cite{guo2019pfld} & 1.167 & 0.051 &$\backslash$ &$\backslash$ \\
        %\cline{}
        ~ &SRT \cite{dong2020srt} & \textcolor{blue}{1.088} & 0.042 &$\backslash$ & $\backslash$\\
        \hline
        \multirow{5}*{Holes} &LD-net & \textcolor{blue}{1.066}  & 0.042 & \textcolor{red}{99.0\%} & 0.41\% \\
        %\cline{}
        ~ &SBR \cite{dong2018sbr}  & 87.529 & 0.015&$\backslash$ &$\backslash$ \\
        %\cline{}
        ~ &AWing \cite{wang2019adaptive} & 14.066 & 0.097 & $\backslash$&$\backslash$ \\
        %\cline{}
        ~ &PFLD \cite{guo2019pfld} & 1.868 & 0.027 & $\backslash$&$\backslash$ \\
        %\cline{}
        ~ &SRT \cite{dong2020srt} & \textcolor{red}{0.947} & 0.037 &$\backslash$ &$\backslash$ \\
    \end{tabular}}
    \caption{Comparison with the existing method for facial landmark detection on the dataset of partial faces and the dataset of faces with holes. We also list the standard deviation while testing. The best result is shown in red, second best in blue.}
    \label{Detect_compare}
\end{table}

In table \ref{Detect_compare}, we compare our method with other facial landmark detection networks. All the networks are trained on partial faces and face with holes, and all the compared networks have no mask output. The testing set consists of 5000 images, and the images are not contained in the training set. 

As shown in table \ref{Detect_compare}, our model outperforms all other methods. SRT and PFLD can also achieve comparable loss in landmark loss but have no mask output for the prominent points' detection. SBR and AWing failed in this sort of dataset.

\subsection{Registration}

For the registration process, we use the same dataset as we used for detection. In the training of the registration network, we set a face with no missing part as a template and register all the input faces to the template by a generated quasi-conformal mapping. Since the quasi-conformal mapping is bijective and we can easily get the inverse mapping based on the output quasi-conformal mapping. 

In Fig. \ref{loss}, we demonstrate the training loss of the registration network. We set the constraint of value $\mu$ at a low level while starting training and adding up its parameter gradually. It is easy to see that value of $\mu$ will drop close to zero at the compensation of a little bit higher rigid error.

\begin{figure}[t!]
\centering
\includegraphics[width=\linewidth]{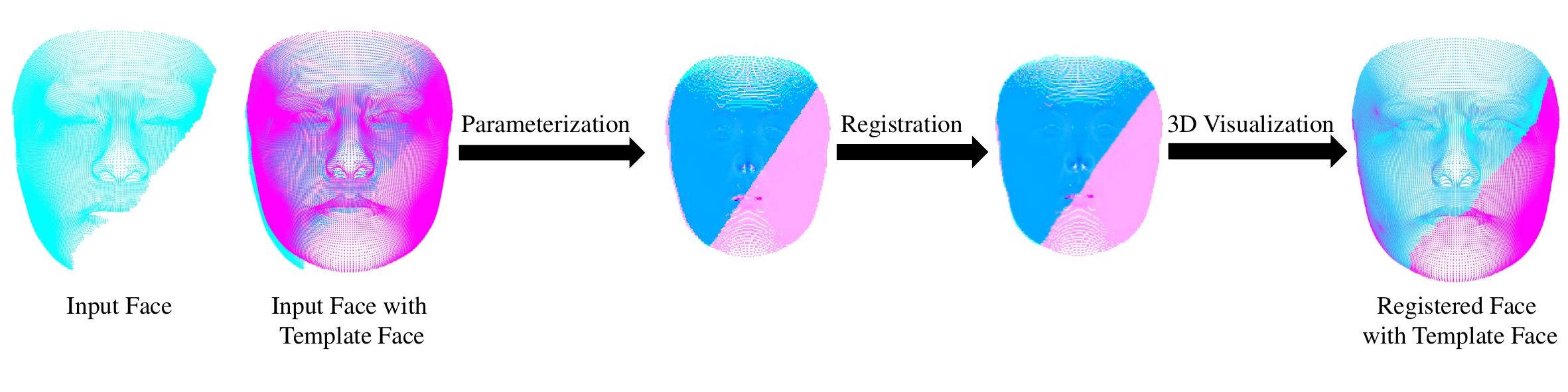}
\includegraphics[width=\linewidth]{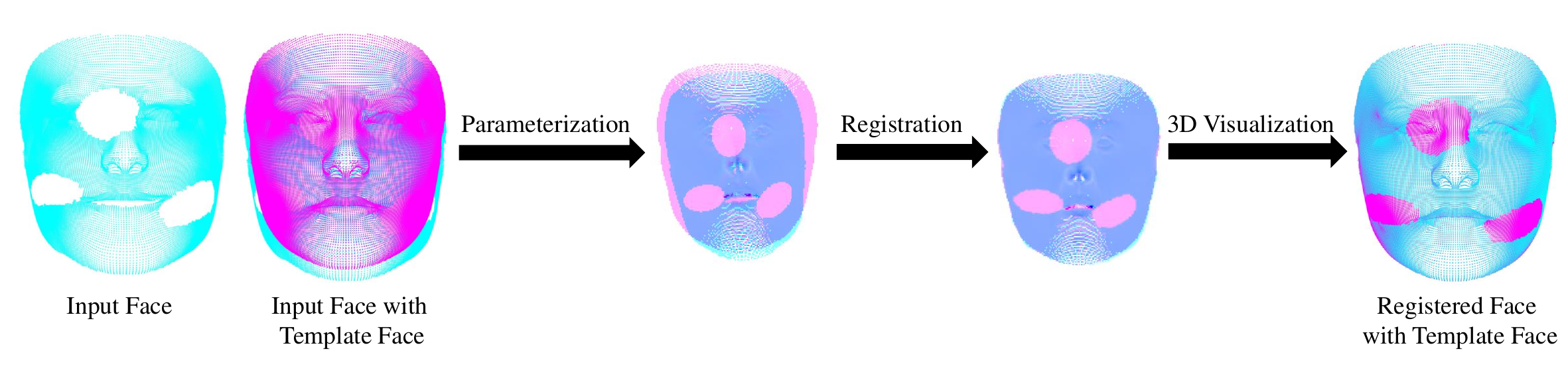}
\caption{On the first column, the input partial face is in pink and the template face is in cyan. The overlapping part is in pink. The second column is the registered partial face with the template face. It can be viewed that the facial features can be aligned with the template. The third column is the mapping generated by our model. }
\label{sin-reg}
\end{figure}

In Fig. \ref{sin-reg}, we demonstrate our model is able to register a partial face to the template full face and have all landmarks aligned.

Since every partial face can be registered to the template and the generated mapping is quasi-conformal, we can get its inverse mapping easily. Thus, for every input partial face, after being registered by our model, we can easily analyze the overlapping features for recognition. In Fig. \ref{reg}, we demonstrate two different partial faces which are registered together.

\begin{figure}[t!]
\centering
\includegraphics[width=\linewidth]{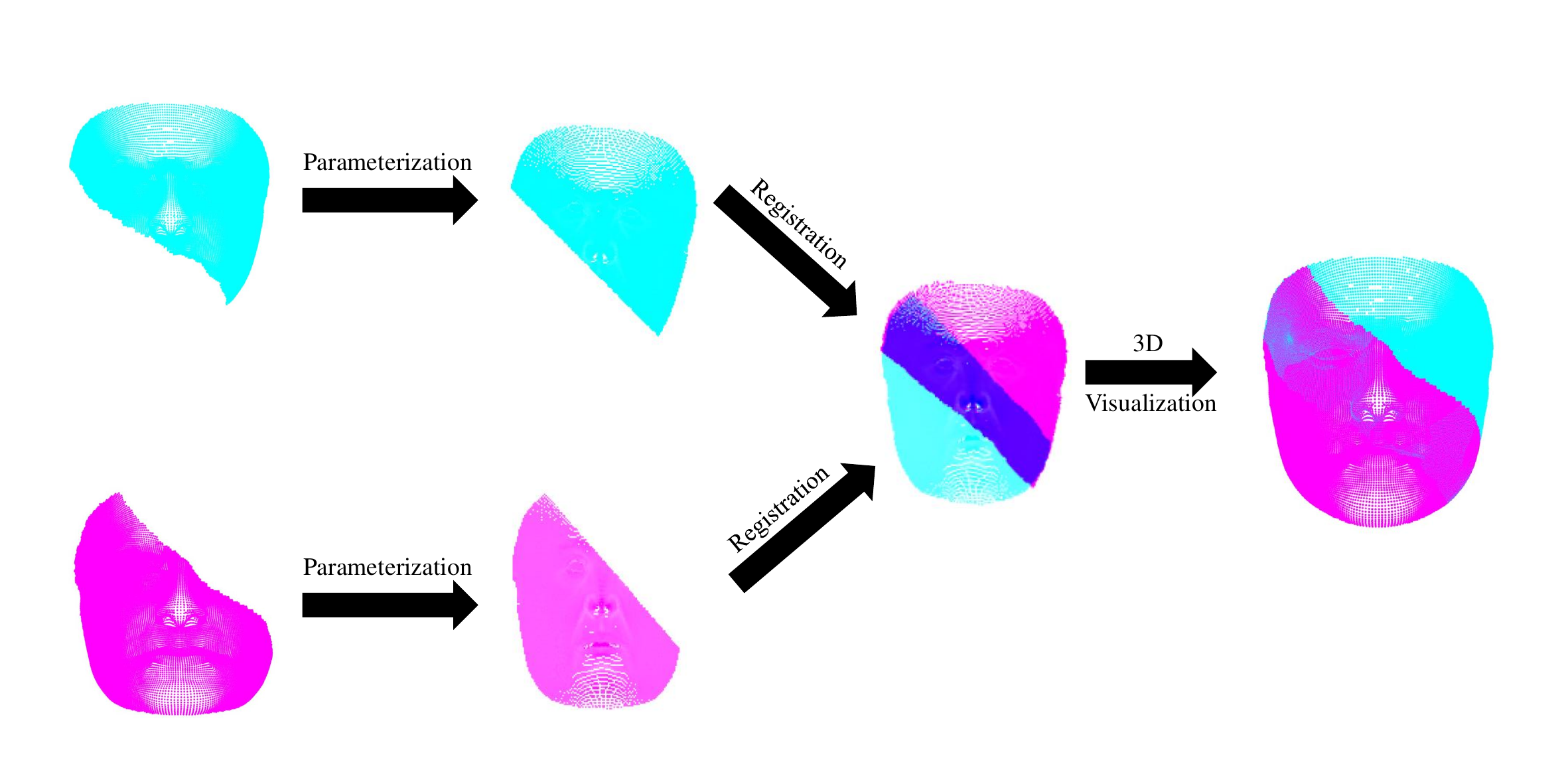}
\caption{Demonstration of two faces with few overlapping. }
\label{few_overlap}
\end{figure}

In Fig. \ref{few_overlap}, we demonstrate that if two faces with very few overlapping, we can also register both faces and easy to analyze the overlapping features for recognition. We will test our model for recognition in the next part.

\subsection{Recognition}

\textbf{Dataset.} We utilize the Labeled Faces in the Wild (LFW) dataset, which comprises 13,233 facial images from 5749 identities. The dataset includes a diverse collection of real-world facial images captured under various conditions, and we select the ones that are easier for 3DMM to reconstruct the 3D face meshes. This enables comprehensive evaluations of our approach. This labeled large dataset allows us to assess the performance and robustness of our method effectively.

%We use the Labeled Faces in the Wild(LFW)\cite{LFWTech} dataset for face recognition.

The data we use for recognition is Labeled Faces in the Wild(LFW)\cite{LFWTech} and we mainly compare with the non-rigid ICP\cite{amberg2007optimal} and the result is shown in \ref{Classify}. To be precise, our method is trained on the previous dataset(use the dataset name) so the LFW dataset is unseen for our model. In Table \ref{Classify}, we can achieve the recognition accuracy of 91.73\%. It is acceptable that it is not close to 100\% since the overlap landmarks could reduce to just a few points. Besides testing the recognition accuracy, we also perform statistical analyses of the result. Suppose $x_0$ is a vector that contains the distances of the same person and $x_1$ contains the distances between different persons. Our hypothesis is that $H_0$: the mean of $x_0$ equals the mean of $x_1$, $H_1$: the mean of $x_0$ less than the mean of $x_1$. In Table \ref{Classify}, the p-value of our output is $3.59 \times 10^{-13}$ and it can be fully convinced that our method could recognize different people reliably.

\begin{table}[]
    \centering
    \resizebox{0.7\textwidth}{11mm}{\begin{tabular}{c|c|c|c}
        Method & Recognition Accuracy & T-test & Time \\
        \hline
        Our Method & 91.73\% & $3.59 \times 10^{-13}$ & 0.102\\
        \hline
        Non-rigid ICP($\epsilon$=1) & 51.24\% &  0.8244 & 116.719\\
        \hline
        Non-rigid ICP($\epsilon$=0.1) & 51.65\% & 0.0997  & 315.766\\
        \hline
        Non-rigid ICP($\epsilon$=0.01) & 52.07\% & 0.6888  & $1.56\times 10^{3}$\\
        %\hline
        %Non-rigid ICP($\epsilon$=0.001) & 42\% & 0.2274  & $9.27\times 10^{3}$\\
    \end{tabular}}
    \caption{Comparison with non-rigid ICP with different parameters for recognition. Our network's performance is far better than non-rigid ICP with any parameter. The t-test shows that the distance of the same person is significantly lower than the distance of the different person after being processed by our model.}
    \label{Classify}
\end{table}

\section{Conclusion}

In conclusion, this work emphasizes the importance of 3D facial analysis and introduced effective techniques for analyzing and comparing partial faces. By using quasi-conformal theory, we achieved accurate parameterization of 3D partial face meshes with minimal distortion. Our study introduced an automated method for detecting landmarks and prominent points on partial faces, eliminating the need for manual annotation. This allowed for precise alignment of partial faces, enabling comprehensive analysis. By generating Beltrami coefficients and applying quasi-conformal mapping, we ensured accurate comparisons between partial faces while preserving bijectivity. The derived distribution of optimal Beltrami coefficients can be used for metric learning in partial face registration problems. Extensive experiments have demonstrated the effectiveness of our approach to landmark detection and partial face registration. Overall, this research contributes to advancing 3D facial analysis by providing efficient techniques for parameterization, landmark detection, and registration of partial faces. These findings have potential implications in various fields, such as biometrics, computer vision, and human-computer interaction. Further research can build upon these methods to improve the accuracy and efficiency of facial analysis.

\section*{Acknowledgement} This work is partially supported by HKRGC GRF (Project ID: 14306721).
%\section{Discussion and Conclusion}

\bibliographystyle{plain}
\bibliography{arxiv}
\end{document}